\pdfoutput=1
\documentclass[11pt]{article}

\usepackage[]{acl}

\usepackage{subcaption}
\usepackage{times}
\usepackage{latexsym}
\usepackage{multicol, multirow}
\usepackage{graphicx}
\usepackage[T1]{fontenc}
\usepackage[utf8]{inputenc}

\usepackage{microtype}

\usepackage{inconsolata}

\usepackage{booktabs}
\usepackage{arydshln} 
\usepackage{enumitem}

\usepackage{paralist, tabularx}

\usepackage{caption} 
\usepackage{upgreek}
\usepackage{float}
\usepackage{array}
\usepackage{colortbl, seqsplit}
\usepackage{lipsum, color}
\usepackage{soul} 
\usepackage{xcolor} 
\usepackage{amsmath}
\usepackage{amssymb}
\usepackage{amsmath}
\usepackage{pifont}
\usepackage{tcolorbox}
\tcbuselibrary{skins}

\colorlet{semitransparentyellow}{yellow!50!white} % 黃色與白色混合，透明度為50%
\newcommand{\hlyellow}[1]{\sethlcolor{semitransparentyellow}\hl{#1}}

\title{Rehearsing Answers to Probable Questions with Perspective-Taking}

\author{Yung-Yu Shih,\textsuperscript{1} Ziwei Xu,\textsuperscript{2} Hiroya Takamura,\textsuperscript{2} Yun-Nung Chen,\textsuperscript{1} Chung-Chi Chen\textsuperscript{2}
\\
\textsuperscript{1} National Taiwan University, Taiwan \\
 \textsuperscript{2} AIST, Japan \\
   \texttt{r12944007@ntu.edu.tw,\{xu.ziwei,takamura.hiroya\}@aist.go.jp}\\ \texttt{y.v.chen@ieee.org, c.c.chen@acm.org}\\
}

\begin{document}
\maketitle
\begin{abstract}
Question answering (QA) has been a long-standing focus in the NLP field, predominantly addressing reading comprehension and common sense QA. However, scenarios involving the preparation of answers to probable questions during professional oral presentations remain underexplored. In this paper, we pioneer the examination of this crucial yet overlooked topic by utilizing real-world QA conversation transcripts between company managers and professional analysts. We explore the proposed task using three causal knowledge graphs (KGs) and three large language models (LLMs). This work provides foundational insights into the application of LLMs in professional QA scenarios, highlighting the importance of causal KGs and perspective-taking in generating effective responses.
\end{abstract}

\section{Introduction}
\label{sec:introduction}

Question answering (QA) is a long-term topic in the natural language processing (NLP) field. Most studies focused on the scenes of reading comprehension and common sense QA~\cite{pal-etal-2023-multitabqa,wang-etal-2023-elaboration,zhang-etal-2023-fc,shaier-etal-2024-desiderata,chen-etal-2024-shot}, but neglect the scenarios of preparing answers to probable questions during a professional oral presentation. 
For example, after the oral presentation at a conference, the presenter should answer questions from the audience; after the press conference, government officials would answer questions from journalists; after a public company's teleconference, the company managers need to respond persuasively to questions from professional analysts. 
Although all the above instances are common scenarios in our daily lives, the discussion on how to assist presenters in handling questions during a presentation is still in the early stages. 
This reveals a significant gap in traditional QA approaches, which may not adequately prepare the specificity required during such professional interactions. 
This paper provides pilot explorations on this important but missing topic by adopting real-world QA conversation transcripts between company managers and professional analysts. 

\begin{table}[t]
    \centering
    \small
    \begin{tabular}{@{}p{7.5cm}@{}}
    \toprule
    \textbf{Question} \\
    Do you think the timeframe for getting Forever 21 \textbf{EBITDA} positive will be similar to that of Aero? \\
    % \addlinespace
    \midrule
    % \addlinespace
    \textbf{Answer 1} \\
    I would say that -- it's a good question. And I would say it's a little more complicated in a little bigger business. And it depends on whether one or two of my guys are going to spend all this time in Los Angeles. So, I'm negotiating it right now, Linda. Stay tuned.  \\
    \midrule
    \textbf{Answer 2} \\
    Based on last quarter's \textbf{sales} increase in North America, I believe that it would be a similar trend.  \\
    \bottomrule
    \end{tabular}
    \caption{Example of Question and Answers. EBITDA denotes ``Earnings Before Interest Taxes Depreciation and Amortization''.}
    \label{table:raw_example}
\end{table}

When rehearsing answers to probable questions, the priority requirement is concreteness. In this work, we highlight that answering with key themes, i.e., the right topics, is the most important metric for evaluating the generated answers. As illustrated in Table~\ref{table:raw_example}, \textbf{Answer 1} is a vague reply, but \textbf{Answer 2} provides a \emph{concrete} answer from the \textbf{sales} aspect, which has a causal relationship to the \textbf{EBITDA} mentioned in the question. 
We have meticulously selected and prepared appropriate datasets to support our methodology. 
Following this line of thought, we adopt causal knowledge graphs (KGs) to guide large language models (LLMs) in rehearsing answers to the given question. 
To evaluate the results from this aspect, we explore three automatic evaluation metrics with human evaluation and compare the correlation coefficient among these metrics with human evaluation results. 
Our study demonstrates how different answers, though equally valid, can significantly differ in their persuasiveness, crucial in financial discussions. We introduce innovative metrics to measure the abstractness of answers and conduct extensive qualitative analyses to compare machine-generated responses with human responses.

In the academic scene and traditional question-answering tasks, \emph{concretely answering questions} is the primary goal. 
% However, choosing the metrics professional analysts care about is another goal and requirement of the QA conversation between company managers and professional analysts. 
Moreover, identifying the metrics valued by professional analysts is essential in QA interactions between managers and analysts.
That is, the ability of perspective-taking, the act of perceiving the expectation from an alternative point of view, is also important in rehearsing answers to professional analysts' questions. 
To this end, in addition to exploring the proposed task with managers' causal KGs, we construct a causal KG with professional analysts' reports, referred to as KG-AR, to activate LLMs' perspective-taking abilities. 
Surprisingly, the employment of this tailored casual KG notably boost performance in select LLMs such as Gemini and GPT-3.5, demonstrating its robust effectiveness. This enhancement illustrates the potential of integrating specialized KGs to enhance the reasoning capabilities of language models.
% using the proposed causal KG leads to better performance regardless of the LLM used. 

To sum up, this paper provides pilot discussions on the following research questions:
\vspace{2mm}

\noindent \textbf{(RQ1)}: To what extent can LLMs assist company managers in rehearsing answers to probable questions?

\vspace{2mm}
    
\noindent \textbf{(RQ2)}: Can we guide LLMs to answer questions from perspective-taking notions with the tailor-made causal KG?

\vspace{2mm}

\noindent \textbf{(RQ3)}: Can causal KGs guide LLMs to learn from historical experience and generate more concrete and informative answers? How can these aspects be empirically validated?

\section{Related Work}

Knowledge graphs, repositories of vast factual content, provide a systematic method for representing knowledge \cite{ji2021survey}. Recent studies indicate that KGs can enhance language modeling by providing background knowledge \cite{renlatent, zhang2019ernie, liu2020k}. The KAPING approach \cite{baek2023knowledgeaugmented} employs KGs to extract relevant triples, corresponding to the input question, with the expectation that directly feeding them into LLMs is beneficial, despite the presence of noise. Our work builds on these developments by proposing the use of KGs to enhance the factual accuracy of LLM responses. By retrieving relevant factual knowledge from KGs, we aim to improve the quality of LLM-generated content without the substantial computational costs associated with retraining. This approach leverages the strengths of pre-trained LLMs while addressing their limitations in handling additional information.

\section{Dataset}
\subsection{Question Answer Pairs}
The transcript of earnings conference calls, a regular meeting between the company's managers and professional analysts, has attracted much attention~\cite{qin-yang-2019-say,mukherjee-etal-2022-ectsum,koval-etal-2023-forecasting}. After managers present their prepared remarks, a QA session opens for analysts to ask questions. Our dataset is collected from Reuters and comprises 205 transcriptions of earnings conference calls sourced from 179 different companies. The task is designed as a model that needs to generate an answer based on the given question. As mentioned in Section~\ref{sec:introduction}, we mainly focus on whether the target financial terms have been correctly generated, as selecting a correct topic is the first step in forming answers.

To sort out financial terms in the questions and answers, we use two financial dictionaries: Investopedia Financial Terms Dictionary and Financial Times Lexicon.\footnote{Please refer to Appendix \ref{sec:Dictionary} for more details.}
These comprehensive resources provided a total of 4,824 unique financial terms. Finally, we obtained 319 QA pairs that contain at least one financial term in the question and at least one financial term in the answer for evaluation.

\subsection{Knowledge Graphs}
In this paper, we adopt two causal knowledge graphs from previous work~\cite{xu2024framework}. The first is FinCaKG-FR, in which the causal relationships between financial terms were extracted from the company's annual report. The second is FinCaKG-ECT, in which the causal relationships were extracted from managers' presentations in the earnings conference call. To explore the perspective-setting, we construct another causal KG, named KG-AR, using the FinArg-1~\cite{chen2023overview} dataset, which includes labels for the causal relationships between financial terms from analysts' reports.
These KGs will guide LLMs with a perspective view through the casual relationship while generating answers, for example, the relationship between EBITDA and sales was pointed out for the  the generation of Answer 2 in Table~\ref{table:raw_example}.
Table~\ref{tab:KGs} provides a brief overview of the three knowledge graphs used in this study, including their scale and complexity.

\begin{table}[t]
\centering
  \resizebox{\columnwidth}{!}{
    \begin{tabular}{l|rrr}
    \toprule
    \textbf{} & \textbf{KG-AR} & \textbf{FinCaKG-FR} & \textbf{FinCaKG-ECT} \\
    \midrule
    \text{Entities} & 4,824 & 1,717 & 546\\ 
    \text{Relations} & 41,007 & 11,633 & 1,802 \\  
    \bottomrule
    \end{tabular}
}
\caption{Comparative overview of KGs.}
\label{tab:KGs}
\end{table}

\section{Evaluation Paradigm}

\subsection{Approaches}
To generate the answers, we adopt three LLMs as baselines: GPT-3.5, Gemini Pro~\cite{team2023gemini}, and LLaMA-3 8B~\cite{llama3modelcard}. 
The input for the baseline models is the manager's presentation and the question from the analyst. \footnote{Please refer to Appendix \ref{app:Generating Prompt} for the prompt.}
We further use the KAPING approach~\cite{baek2023knowledgeaugmented} for employing KG to extract relevant triples. These triples correspond to the LLM's input. That is, we enhance the LLM's capabilities by integrating relevant facts, specifically tailored to the input question. These facts are pre-appended to the original question to form an enriched prompt, which is then fed into the LLMs for answer generation. Our framework operates entirely in a zero-shot manner, eliminating the need for additional model training and aiming to bridge the gap between the LLM’s pre-stored knowledge and the factual accuracy required for domain-specific answer generation.

\subsection{Evaluation Metrics}
Our metrics evaluate the relevance and richness of terms in generated answers (\(GA\)) relative to the reference answer (\(RA\)) by examining three key aspects.
Each \(GA\) includes a list of financial terms \(GA_{\text{Terms}}\), and similarly, each \(RA\) also contains a list of financial terms \(RA_{\text{Terms}}\). The list of financial terms appearing in the question (\(Q\)) is denoted as \(Q_{\text{Terms}}\). We analyze the relationships between these term sets by examining their intersections and differences, providing insight into both the straightforward relevance and inferential depth of the responses.

\paragraph{Accurate Term Ratio (ATR)} quantifies the uniqueness of terms in the generated answers compared to the ground truth:
\begin{equation}
\text{ATR} = \frac{\left|GA_{\text{Terms}}\right|}{\left|RA_{\text{Terms}}\right|}
\end{equation}

\paragraph{Accurate Copy Ratio (ACR)} measures the overlap of unique terms between the question and the answers, assessing direct relevance:

\begin{equation}
\text{QAR} = \frac{\left|GA_{\text{terms}} \cap Q_{\text{terms}}\right|}{\left|RA_{\text{terms}} \cap Q_{\text{terms}}\right|} 
\end{equation}

\paragraph{Inferential Insight Ratio (IIR)} is determined by the proportion of unique terms in the answers that do not appear in the question, offering insights into the depth of the generated content:
\begin{equation}
\text{IIR} = \frac{\left|GA_{\text{Terms}} \setminus Q_{\text{Terms}}\right|}{\left|RA_{\text{Terms}} \setminus Q_{\text{Terms}}\right|}
\end{equation}

In addition to the aforementioned term-based metrics, we conduct a human evaluation based on the following two aspects, with scores ranging from 0 to 10.

\paragraph{Information Richness (INFO)} measures the extent and breadth of information in a response. Responses that offer a comprehensive background, extensive analysis, and cover a wide range of ideas are rated highly. Medium scores go to responses with fair information but limited in some areas, and low scores are given to those with minimal and superficial information.

\paragraph{Concreteness (CON)} evaluates the level of detail in an answer. High scores are reserved for responses that provide detailed numerical data and empirical evidence, medium scores for those with some specifics, and low scores for vague answers lacking detailed examples.

\begin{table}[t]
  \centering
  \resizebox{\columnwidth}{!}{
    \begin{tabular}{ll|ccc}
    \toprule
    \multicolumn{1}{c}{\bf LLM} & \multicolumn{1}{c|}{\bf KG} & \multicolumn{1}{c}{\bf ATR} & \multicolumn{1}{c}{\bf ACR} & \multicolumn{1}{c}{\bf IIR} \\
    \midrule
    \multirow{4}[2]{*}{GPT-3.5} & \multicolumn{1}{c|}{-} & 22.35 & \underline{\textbf{45.41}} & 14.60 \\
          & FinCaKG-FR & 21.62 & 42.42 & 16.00 \\
          & FinCaKG-ECT & 22.38 & 34.12 & 16.34 \\
          & KG-AR (PT) & \underline{\textbf{24.06}} & 38.69 & \underline{\textbf{18.54}} \\
    \midrule
    \multirow{4}[2]{*}{Gemini Pro} & \multicolumn{1}{c|}{-} & 15.47 & 36.61 & 7.32 \\
          & FinCaKG-FR & 11.83 & 26.60 & 6.21 \\
          & FinCaKG-ECT & 13.83 & 25.94 & 8.68 \\
          & KG-AR (PT) & \underline{15.91} & \underline{37.00} & \underline{9.47} \\
    \midrule
    \multirow{4}[1]{*}{LLaMA-3 8B} & \multicolumn{1}{c|}{-} & 19.53 & \underline{21.41} & 17.62 \\
          & FinCaKG-FR & 19.98 & 19.17 & \underline{18.38} \\
          & FinCaKG-ECT & \underline{20.52} & 20.17 & 17.87 \\
          & KG-AR (PT) & 19.08 & 18.62 & 17.31 \\
    \bottomrule
    \end{tabular}%
    }
    \caption{Automatic evaluation (\%). \textbf{Bold} fonts denote the best performance among all settings, and \underline{underlined} ones denote the best performance among different KGs with the same LLM. PT: Perspective-Taking.}
  \label{tab:Automatic}%
\end{table}%

\section{Results and Analysis}

\subsection{KG vs. Rehearsing Answers}
Table~\ref{tab:Automatic} provides the answer to (RQ1) and (RQ2).\footnote{Statistics of other metrics such as ROUGE are reported in Appendix \ref{app:evaluation}.} First, using KGs can lead to higher ATR and IIR regardless of the LLM used. These results show that KGs can help guide LLMs in generating correct financial terms and making further implicit inferences. However, making implicit inferences (IIR) and copying terms directly (ACR) involves a trade-off, and thus, two out of three LLMs achieved higher ACR without using KGs. Second, FinCaKG-FR and FinCaKG-ECT, which are constructed based on the narratives of companies and managers, negatively impacted the ATR of two out of three LLMs (GPT-3.5 and Gemini Pro). In contrast, KG-AR, which is constructed based on the analysis of analysts, obtained the best scores. As mentioned in Section \ref{sec:introduction}, these results indicate that using the audience's causal KGs could inject perspective-taking ability into models. Given that analysts' reports after earnings calls influence market participants' views of the company~\cite{piotroski2004influence} and their reports are based on the information revealed in the earnings calls~\cite{keith-stent-2019-modeling}, our results provide insights into leveraging audience causal KGs to activate the perspective-taking ability of LLM. 

\subsection{Human Evaluation}
To answer (RQ3), we assessed the quality of the generated answers manually, since there are no automatic metrics that can evaluate information richness and concreteness aspects. 
Three annotators, who graduated from financial or economics departments, were paid 36\% higher than the minimum wage stipulated by law. 
The results are shown in Table \ref{tab:Human}. 
First, the scores in the answers generated by the LLM were generally higher than the manager's answers, indicating a satisfactory performance in conveying information. 
Second, with KG-AR, two out of three LLMs achieved higher INFO scores, indicating that using KGs can enhance information richness. 
Third, using KGs did not improve the models' ability to select better numerical evidence to provide concrete answers. These findings suggest that future work can also aim to enhance the concreteness of rehearsed answers.

Drawing inspiration from \citet{chiang2023large}, we also used GPT-4 as an automatic evaluator to check whether future studies can rely on GPT-4 to evaluate the generated answers.\footnote{The prompts are provided in Appendix \ref{sec: LLM_eval_prompt}.} We established the evaluation framework based on previous studies~\cite{zhong-etal-2022-towards,chan2023chateval} and employed Pearson, Spearman, and Kendall correlation coefficients. Table~\ref{tab:Correlation} shows the statistics of correlation coefficients between GPT-4 and human evaluators. Overall, GPT-4's annotations are correlated with those of human evaluators. However, although GPT-4 has a moderate level (> 0.3) of correlation when evaluating from a concreteness aspect, there is still room for improvement in automatic evaluation. Our suggestion based on these findings is that future work can evaluate the results with GPT-4, but manual checking is still needed, given the current correlation coefficient level.

\begin{table}[t]
  \centering
  \small
    \begin{tabular}{lcrr}
    \toprule
    \multicolumn{1}{c}{\bf LLM} & \bf KG-AR & \multicolumn{1}{c}{\bf INFO} & \multicolumn{1}{c}{\bf CON} \\
    \midrule
    \multicolumn{1}{c}{-} & - & 5.15 & 5.50 \\
    \midrule
    \multirow{2}[2]{*}{GPT-3.5} & w/o   & \underline{\textbf{6.26}} & \underline{\textbf{6.24}} \\
          & w/    & 6.00  & 5.97 \\
    \midrule
    \multirow{2}[2]{*}{Gemini Pro} & w/o   & 5.16  & \underline{5.87} \\
          & w/    & \underline{5.42} & 5.34 \\
    \midrule
    \multirow{2}[1]{*}{LLaMA-3 8B} & w/o   & 6.16  & \underline{6.13} \\
          & w/    & \underline{6.37} & 6.08 \\
    \bottomrule
    \end{tabular}%
  \caption{Human evaluation.The first row displays the baseline scores of managers' answers.}
  \label{tab:Human}%
\end{table}%

\begin{table}[t]
  \centering
  \small
    \begin{tabular}{lrrr}
    \toprule
          & \multicolumn{1}{c}{\bf Pearson} & \multicolumn{1}{c}{\bf Spearman}& \multicolumn{1}{c}{\bf Kendall} \\
    \midrule
    INFO  & 0.30  & 0.28  & 0.23 \\
    CON   & 0.41  & 0.38  & 0.31 \\
    \bottomrule
    \end{tabular}%
  \caption{Correlation between scores evaluated by GPT-4 and human.}
  \label{tab:Correlation}%
  \vspace{-3mm}
\end{table}%

\section{Conclusion}
We have explored the under-researched domain of preparing answers to probable questions in professional oral presentations. Our investigation focused on the efficacy of using causal KGs and LLMs in rehearsing answers. 
Our findings indicate that KGs significantly enhance the ability of LLMs to generate relevant financial terms. 
Future work can further refine our methods to enhance the concreteness and reliability of rehearsed answers, and ultimately improve the quality of professional presentations and interactions.

\section*{Limitation}
First, due to the limited availability of data, the dataset used in this paper consists exclusively of real-world QA conversation transcripts between company managers and professional analysts. While this dataset is highly relevant for the study's context, it limits the generalizability of the findings to other professional or informal presentation settings. Future studies should consider incorporating diverse datasets to enhance the applicability of the results across different domains.
Second, the automatic evaluation metrics employed, such as ATR and IIR, might not fully capture the nuanced quality of the generated answers. Although human evaluation was conducted to assess information richness and concreteness, the reliance on subjective judgment can introduce bias. Developing more comprehensive and objective evaluation metrics could provide a more accurate assessment of the answer quality.

\bibliography{custom}

\appendix

\section{Dictionary}
\label{sec:Dictionary}
We accessed 
Investopedia Financial Terms Dictionary\footnote{\href{https://www.investopedia.com/financial-term-dictionary-4769738}{https://www.investopedia.com/financial-term-dictionary-4769738}} and Financial Times Lexicon\footnote{\href{https://markets.ft.com/glossary/searchLetter.asp?letter=C}{https://markets.ft.com/glossary/searchLetter.asp?letter=C}} on Jan 20, 2024.

\section{Generating Answer Prompts}
\label{app:Generating Prompt}
\subsection{No presentation and no knowledge graph}
\begin{tcolorbox}[width=\columnwidth,colback=white]
\small
\begin{verbatim}
Below are the questions during the earnings 
call: {Question}

Assume the role of a company executive, and 
provide an oral response in English narrative 
format, without paragraphs:
\end{verbatim}
\end{tcolorbox}

\subsection{Presentation without knowledge graph}
\begin{tcolorbox}[width=\columnwidth,colback=white] % start here
\small
\begin{verbatim}
Below are the questions during the earnings 
call: {Question}

Before the earnings call, the executive 
prepared remarks: {Prepare Remarks}

Assume the role of a company executive and 
provide an oral response in English narrative 
format, without paragraphs:
\end{verbatim}
\end{tcolorbox}

\subsection{Knowledge graph without presentation} 
\begin{tcolorbox}[width=\columnwidth,colback=white] % start here
\small
\begin{verbatim}
Below are the questions during the earnings 
call: {Question}
        
Please infer based on the relevant terms 
obtained from the earnings call knowledge 
graph: {Extract related entities from KG}

Assume the role of a company executive and 
provide an oral response in English narrative 
format, without paragraphs:
\end{verbatim}
\end{tcolorbox}

\subsection{Both presentation and knowledge graph included}
\begin{tcolorbox}[width=\columnwidth,colback=white] % start here
\small
\begin{verbatim}
Below are the questions during the earnings 
call: {Question}

Before the earnings call, the executive 
prepared remarks: {Prepare Remarks}

Please infer based on the relevant terms 
obtained from the earnings call knowledge 
graph: {Extract related entities from KG}

Assume the role of a company executive and 
provide an oral response in English narrative 
format, without paragraphs:
\end{verbatim}
\end{tcolorbox}

\section{Other Metrics}
\label{app:evaluation}
We report traditional metrics, ROUGE and BERTScore~\cite{zhang2019bertscore} in Table~\ref{tab:other metrics}.

\section{LLM Evaluation Prompts}
\label{sec: LLM_eval_prompt}
We list the prompt we use in this section.
In the main content of the paper and in the following parts, we use different highlight colors to represent different parts of the prompt.
A prompt is composed of four parts: 
(1) the \textbf{{descriptions of the rating task}}, (2) the \textbf{{definition and rating criteria}} of the attribute (\textit{concreteness}, \textit{information richness}, and \textit{relevance and accuracy}) to be rated, and (3) \textbf{{a sentence used to prompt the LLM to give the rating}}.

\begin{table}[t]
  \centering
  \resizebox{\columnwidth}{!}{
    \begin{tabular}{ll|rrr|r}
    \multicolumn{1}{c}{\multirow{2}[1]{*}{LLM}} & \multicolumn{1}{c|}{\multirow{2}[1]{*}{KG}} & \multicolumn{3}{c|}{ROUGE} & \multicolumn{1}{c}{\multirow{2}[1]{*}{BERTScore}} \\
          &       & \multicolumn{1}{c}{1} & \multicolumn{1}{c}{2} & \multicolumn{1}{c|}{L} &  \\
    \hline
    \multirow{4}[2]{*}{GPT-3.5} & \multicolumn{1}{c|}{-} & 29.98 & 5.12  & 14.89 & 83.46 \\
          & FinCaKG-FR & 27.90 & 4.43  & 14.00 & 83.10 \\
          & FinCaKG-ECT & 28.03 & 4.48  & 14.00 & 83.13 \\
          & KG-AR & 27.85 & 4.28  & 13.86 & 82.97 \\
    \hline
    \multirow{4}[2]{*}{Gemini Pro} & \multicolumn{1}{c|}{-} & 24.49 & 4.57  & 12.63 & 83.01 \\
          & FinCaKG-FR & 21.01 & 3.69  & 11.53 & 82.67 \\
          & FinCaKG-ECT & 22.28 & 3.84  & 12.02 & 82.55 \\
          & KG-AR & 24.25 & 4.55  & 12.36 & 82.61 \\
    \hline
    \multirow{4}[1]{*}{LLaMA-3 8B} & \multicolumn{1}{c|}{-} & 25.67 & 3.68  & 12.04 & 81.54 \\
          & FinCaKG-FR & 27.51 & 3.82  & 12.98 & 81.77 \\
          & FinCaKG-ECT & 27.93 & 3.96  & 12.98 & 81.82 \\
          & KG-AR & 28.28 & 4.18  & 13.30 & 81.81 \\
    \end{tabular}%
    }
  \caption{Evaluation --- other metrics.}
  \label{tab:other metrics}%
\end{table}%

\begin{tcolorbox}[width=\columnwidth,colback=white]
\small
\begin{verbatim}
Task: Evaluate an answer to the question
using the following criteria. 

Question Asked: {question} 

Answer Provided: {answer}

1. Concreteness (1-10): Assess how specific
and detailed the answer is.
    
- 1: Extremely vague, lacking specific details 
or examples. - 5: Provides moderate detail and 
examples but could be more specific. - 10: 
Exceptionally detailed, with clear, specific 
examples or data supporting the points.
    
2. Information Richness (1-10): Gauge the 
breadth of the information provided.
- 1: Offers minimal useful information, 
background, or analysis. - 5: Contains a fair 
amount of information and analysis but lacks 
breath in crucial areas. - 10: Highly 
informative, with comprehensive background, 
detailed analysis, and insightful commentary.

3. Relevance and Accuracy (1-10): Determine 
if the answer directly and accurately 
addresses the question. - 1: Does not 
address the question, lacking relevance 
and accuracy. - 5: Partially addresses 
the question but misses full relevance 
or contains inaccuracies. - 10: Directly 
and accurately addresses the question 
in full, with high relevance and accuracy.

Score each criterion based on how well the 
answer meets these benchmarks, adjusting 
scores between 1, 5, and 10 as needed.

\end{verbatim}
\end{tcolorbox}

\section{Case Study}
\subsection{Discussing Concreteness}
 Table~\ref{table:raw_example10} compares the concreteness of answers by contrasting the original responses with those generated by LLM. The initial response, derived from a dataset, is vague and missing detailed financial indicators. In contrast, the response enhanced by knowledge graph data is expected to be more detailed, showcasing the importance of structured knowledge in delivering clearer and more actionable insights in financial analysis.

\begin{table}[hbt!]
    \centering
    \small
    \begin{tabular}{@{}p{8cm}@{}}
    \toprule
    \textbf{Question} \\
    Are you much higher than where you were in '19? Or is it kind of adjusted for occupancy changes, like margins more or less the same in your mind or kind of expense ratios? \\
    \addlinespace[3pt]
    \midrule 

    \textbf{Non-specific Answer from Dataset}\\
    I'd say -- well, other than the drop in occupancy, I think in terms of operating, it's probably pretty similar to '19. Yes.\\
    \addlinespace[3pt]
    \noalign{\vskip 0.5ex} 
    \cdashline{1-1}
    \noalign{\vskip 0.75ex}

    \textbf{Answer After Integrating Knowledge Graph}\\
    TIn comparing our current performance to 2019, it's important to note that our cash flow has increased significantly to nearly \$3 billion year-to-date, consistent with pre-pandemic levels. We have seen growth in leasing volumes, occupancy rates, shopper traffic, and retail sales. Our domestic property NOI increased by 24.5\% year over year for the quarter and 8.8\% year to date, showcasing positive momentum.

    Occupancy stands at 92.8\%, with average base rent at \$53.91. Mall sales have rebounded, up 11\% compared to 2019 for the third quarter and an impressive 43\% year over year. This performance is noteworthy, especially considering the challenges posed by various COVID restrictions. We have diversified revenue streams with investments in different platforms like SPARC, Penney, and ABG, all of which are performing well.
    
    In terms of guidance, we raised our expectations from \$10.70 to \$10.80 last quarter to \$11.55 to \$11.65 per share, indicating strong growth. Our dividend has also seen a significant increase, with the latest announcement of a \$1.65 per share dividend for the fourth quarter, representing a 10\% sequential increase and 27\% year over year growth.
    
    Overall, we are pleased with the progress we have made and are confident in our ability to continue delivering value to our shareholders.\\
    
    \addlinespace[3pt]
    
    \bottomrule
    \end{tabular}
    \caption{Case Study.}
    \label{table:raw_example10}
\end{table}

\begin{table}[t!]
    \centering
    \small
    \begin{tabular}{@{}p{7.5cm}@{}}
    \toprule
    \textbf{Question} \\
    Good morning. So, understood on the difficulty of giving formal guidance at this point, but could we just maybe go through some more granularity around how you're seeing some of these different issues, impact to businesses? And I think probably, I want to focus mostly on ROVs, ... \\
    \addlinespace[3pt]
    % \addlinespace
    \midrule 
    % \addlinespace
    % \textbf{Answer} \\
    % \midrule

    \textbf{Generated Output with FinCaKG-FR}\\
    Yes, thank you for the insightful question regarding our free \hlyellow{cash flow} \hlyellow{outlook} for the year. We have indeed been closely monitoring our \hlyellow{net cash} position, especially given the challenging \hlyellow{market conditions} we have faced in the first \hlyellow{quarter}. 
    While we did experience a decrease in our \hlyellow{cash balance} primarily due to ... \\
    \addlinespace[3pt]
    \noalign{\vskip 0.5ex} 
    \cdashline{1-1}
    \noalign{\vskip 0.75ex}

    \textbf{Generated Output with FinCaKG-ECT}\\
    Certainly. Based on the information provided during the earnings call, it is evident that Oceaneering faced significant \hlyellow{challenges} in the first \hlyellow{quarter} of 2020 due to the impact of the COVID-19 pandemic and the \hlyellow{volatility} in the \hlyellow{crude oil} \hlyellow{market}. Despite these challenges, we were able to exceed expectations with adjusted \hlyellow{net income} of \$3.5 million, driven by better-than-anticipated performance in our energy-focused businesses. As we ...\\
    \addlinespace[3pt]
    \noalign{\vskip 0.5ex} 
    \cdashline{1-1}
    \noalign{\vskip 0.75ex}
    
    \textbf{Generated Output with KG-AR}\\
    Yes, it is indeed a fair assumption that we should see a significant reversal of \hlyellow{cash movement} in Q2 compared to Q1. We have been working on implementing \hlyellow{cost} \hlyellow{reduction} measures and \hlyellow{efficiency} improvements across all segments, which should positively impact our free \hlyellow{cash flow} in the \hlyellow{second} \hlyellow{quarter}. 
    Despite the challenging \hlyellow{global} \hlyellow{environment}, we have exceeded expectations in our first \hlyellow{quarter} \hlyellow{results}, showing the resilience and adaptability of our \hlyellow{business}. 
    We are closely monitoring the evolving \hlyellow{market conditions} and taking decisive \hlyellow{actions} to ... \\
    \addlinespace[3pt]
    
    \bottomrule
    \end{tabular}
    \caption{Case Study. Financial terms are highlighted in yellow. 
    % (Version 2, short version)
    }
    \label{table:raw_example9_2}
\end{table}

However, our results reveal that integrating a knowledge graph does not significantly enhance concreteness. We found that concreteness depends not only on causal reasoning but also on the accuracy of numerical data and other factors. Therefore, evaluating concreteness requires additional information beyond just the content of the presentation.

\subsection{Different Knowledge Graphs}
As illustrated in Table~\ref{table:raw_example9_2}, a comparative analysis of the outcomes from three distinct knowledge graphs provides insight into their effectiveness in addressing specific strategic questions.

FinCaKG-FR focuses closely on financial stability and liquidity preservation, detailing active cash flow management and strategic financial stability measures. This response is more aligned with the financial nuances of the query but still leaves room for deeper exploration of strategic realignment towards opportunities in renewables.

On the other hand, the response generated from FinCaKG-ECT paints a comprehensive picture of a company's strategic adjustments in a challenging environment. It discusses broad cost-cutting measures and efficiency drives essential for navigating economic volatilities influenced by the COVID-19 pandemic and fluctuations in the crude oil market. Even so, while this output offers a broad strategic overview, it lacks a direct response to the nuanced aspects of the query regarding jurisdictional footprint optimization and internal funding mechanisms.

The response from KG-AR, however, stands out by directly addressing the reversal of cash flows expected in Q2, emphasizing cost reduction and efficiency improvements. This response not only confirms the resilience and adaptability of the business under tough market conditions but also aligns closely with strategic considerations of balance sheet optimization and seizing growth opportunities in the renewables sector. The narrative crafted by the KG-AR's response effectively matches the query’s strategic depth, offering a clear and confident reflection on anticipated improvements and strategic directions.

\end{document}